\pdfoutput=1

\documentclass[11pt]{article}

\usepackage{ACL2023}

\usepackage{times}
\usepackage{latexsym}

\usepackage[T1]{fontenc}

\usepackage[utf8]{inputenc}

\usepackage{microtype}

\usepackage{inconsolata}

\usepackage{url}            
\usepackage{booktabs}       
\usepackage{amsfonts}       
\usepackage{nicefrac}       

\usepackage{enumitem}
\usepackage{graphicx} 
\usepackage{subcaption}

\usepackage{amsmath}
\usepackage{bm}
\usepackage{comment}

\newcommand\our{\textsc{ELLA-V}}

\newcommand{\ie}{\textit{i}.\textit{e}.}
\newcommand{\eg}{\textit{e}.\textit{g}.}
\newcommand{\wrt}{\textit{w}.\textit{r}.\textit{t}.}

\title{ELLA-V:  Stable Neural Codec Language Modeling \\ with Alignment-guided Sequence Reordering}

\author{%
    Yakun Song\textsuperscript{1}, 
    Zhuo Chen\textsuperscript{2}, 
    Xiaofei Wang\textsuperscript{2}, 
    Ziyang Ma\textsuperscript{1}, 
    Xie Chen\textsuperscript{1} \\
    \textsuperscript{1}Shanghai Jiao Tong University, Shanghai, China\\
    \textsuperscript{2}Microsoft, One Microsoft Way, Redmond, USA\\
}

\begin{document}

\maketitle

\begin{abstract}
The language model (LM) approach based on acoustic and linguistic prompts, such as VALL-E, has achieved remarkable progress in the field of zero-shot audio generation. However, existing methods still have some limitations: 1) repetitions, transpositions, and omissions in the output synthesized speech due to limited alignment constraints between audio and phoneme tokens; 2) challenges of fine-grained control over the synthesized speech with autoregressive (AR) language model; 3) infinite silence generation due to the nature of AR-based decoding, especially under the greedy strategy. To alleviate these issues, we propose \our\footnote{\our{} is the word ``VALL-E'' spelled backwards, to highlight that our model changes the sequence order of VALL-E input.}, a simple but efficient LM-based zero-shot text-to-speech (TTS) framework, which enables fine-grained control over synthesized audio at the phoneme level. The key to \our{} is interleaving sequences of acoustic and phoneme tokens, where phoneme tokens appear ahead of the corresponding acoustic tokens. The experimental findings reveal that our model outperforms VALL-E in terms of accuracy and delivers more stable results using both greedy and sampling-based decoding strategies. The code of \our{} will be open-sourced after cleanups\footnote{VALL-E is not officially open source. We reproduced it and open-sourced it in this repository.}. Audio samples are available
at \url{https://ereboas.github.io/ELLAV/}.
\end{abstract}

\section{Introduction}

Recently, deep generative AI has achieved remarkable results in various tasks, leading to the emergence of many transformative real-world applications~\citep{gpt3, ramesh2022hierarchical, ddpm, rombach2022high, audiolm, vits, chiang2019cluster}. With the advancement of generative models, there have been rapid developments in the field of speech synthesis as well. In particular, zero-shot TTS technology has gained increasing attention because it can synthesize high-quality target voices without the need of specified speaker's training data.
As a state-of-the-art generative model family, diffusion models~\citep{sohl2015deep, ddpm, song2020improved} progressively add noise to the training data and then learn the reverse process to generate samples. By leveraging diffusion models and their variants~\citep{sohl2015deep, ddpm, song2020improved, song2021scorebased, flowmatching}, many works have successfully applied them to the audio domain~\citep{gradtts, prodiff, makeanaudio, naturalspeech}. 
Another major class of generative models is language modeling based on Transformer~\citep{vaswani2017attention}. \citet{bert, t5, bart} utilize encoder-only or encoder-decoder architectures to build masked language models so that they selectively focus on relevant segments and effectively model relationships in long sequences. However, masked language model often requires fine-tuning to adapt to specific tasks, which can be inconvenient for practical usage and deployment. On the other hand, AR language models use a decoder-only architecture to predict the next token in a sequence as the training objective, which has demonstrated extremely powerful few-shot and zero-shot capabilities in many generative tasks~\citep{gpt3, lamda, palm}. In light of this, VALL-E~\citep{VALLE} and subsequent works~\citep{speartts, audiopalm, wang2023lauragpt} have successfully employed decoder-only language model for zero-shot TTS. 
These approaches first quantize the speech signal into a series of discrete acoustic tokens. Subsequently, they employ an AR language model to predict coarse-grained acoustic tokens, eliminating the necessity for explicit duration predictors or speaker encoders. Once trained on a large-scale corpus, such as LibriLight~\citep{librilight}, these approaches are capable of synthesizing speech with competitive fidelity and naturalness in a zero-shot manner. 

While VALL-E and its variants have achieved numerous impressive milestones, they still possess certain limitations that impact practical deployment. For instance, existing methods~\citep{VALLE,speartts} directly concatenate phoneme tokens and acoustic tokens as a whole sequence to train language models. 
In this way, the alignment between audio and phoneme sequences is completely learned through the self-attention in the transformer, making it potentially unstable as self-attention does not explicitly capture the monotonic alignment between audio and phoneme.  
Additionally, the decoder-only language model architecture can lead to potential attention degradation issues~\citep{fu2023decoder}, where the alignment quality between the target audio sequence and the source phoneme sequence deteriorates as the generated sequence increases, resulting in inaccurate or low-quality speech outputs. 

Another limitation stems from the nature of AR language modeling. Specifically, given a sequence $\mathbf{x}$, the standard AR language model factorizes the likelihood $p(\mathbf{x})$ over the dimensions of $\mathbf{x}$ via the chain rule $p(\mathbf{x})=\prod_{t=0}^{T}p(x_t|\mathbf{x}_{<t})$. AR models predict the current tokens solely based on the historical tokens without users' control in the inference process, and sometimes generate semantic repetitions or incoherence in the generated output~\citep{xlnet, gpt3}.
In the TTS task, correspondingly, VALL-E cannot directly determine which segment of the output audio corresponds to which prompt phoneme, thus there is no trivial way to promptly detect and prevent issues occurring in the generation process. 
These drawbacks can manifest as meaningless phoneme repetitions, transpositions, omissions, or even catastrophic \emph{infinite silence}, \ie, during the process of generation, the model anomalously outputs silence or noise tokens for an extended period of time without stopping. Specifically, Table \ref{tab:intro} demonstrates the word error rate (WER) and the probability of the \emph{infinite silence} in VALL-E samples at different threshold top-$p$ for nuclear sampling~\citep{nuclearsampling}. The detailed experimental setup is described in Section \ref{sec:exp}. Notably, a shift in the decoding strategy of VALL-E from fully sampling-based to fully greedy-based leads to a marked decline in sample quality. It should be emphasized that while sampling-based stochastic decoding strategies have advantages in terms of synthesis diversity, deterministic decoding strategies (\eg, beam search and its variants) are more suitable for cases where there is less tolerance for synthesis errors and more emphasis on fluency and coherence~\citep{ippolito-etal-2019-comparison}.

\begin{table}
  \caption{Comparison of VALL-E’s zero-shot TTS performance across various top-$p$ thresholds in nuclear sampling. INF\% denotes the probability of \emph{infinite silence}, which refers to instances where generation continues without stopping even when its duration exceeds twice the original length.}
  \label{tab:intro}
  \centering
  {\footnotesize
  \begin{tabular}{lll}
    \toprule
    Top-$p$       & WER\%   & INF\%    \\
    
    \midrule
    1             & 5.47   & 0.00    \\
    0.99          & 5.00   & 0.20    \\
    0.95          & 10.99  & 19.06   \\
    0.9           & 20.85  & 41.43   \\
    0.7           & 37.71  & 76.76   \\
    0.4           & 46.59  & 84.39   \\
    0.0 (greedy)  & 49.26  & 87.29   \\
    
    \bottomrule
  \end{tabular}
}
\end{table}

Faced with the pros and cons of the existing methods, we introduce \our, a simple but effective language model approach for zero-shot TTS. \our{} proposes a generalized AR (GAR) language model to generate the first layer of residual vector quantizer (RVQ) codes of a neural codec model. Then as with VALL-E, \our{} employs a non-autoregressive (NAR) language model to obtain codes of the other RVQs.
Our core innovation lies in 3 fold:

\begin{itemize}
  \item Firstly, \our{} inserts phone tokens into the corresponding positions of the acoustic sequence. Unlike existing methods, Connecting phoneme tokens with their corresponding acoustic tokens can help the language model capture the alignment between phoneme and acoustic modalities in local dependencies.
  
  \item Secondly, instead of maximizing the expected log-likelihood of the hybrid sequence under a conventional casual mask or a prefix mask like VALL-E and UniLM~\citep{unilm}, \our{} computes loss only on acoustic tokens and special tokens \verb|EndOfPhone(EOP)| and \verb|EndOfSentence(EOS)|. This training objective not only reduces the redundant computation of cross-modal alignment in the output based on experimental results, but also provides a natural way to have fine-grained control in inference: the model predicts \verb|EOP|, and then the user provides the next phone token. Meanwhile, \our's GAR model always maintains awareness of the phoneme it is currently synthesizing, allowing it to promptly detect and truncate any abnormal phoneme to avoid any possible \emph{infinite silence} issue.

  \item Thirdly, we further propose an improvement to the input sequence. We introduce \emph{local advance}, which involves shifting the \verb|EOP| token and the next-word phoneme token a few frames ahead. Intuitively, the pronunciation of a phoneme, especially its ending, is not only influenced by the context in history but also by the upcoming phonemes. By advancing these special tokens, the GAR model can better utilize local dependencies to predict the pronunciation of the current phoneme.
  
\end{itemize}

Experimental results, using comparable model configurations and 960 hours of speech data from LibriSpeech~\citep{librispeech} as a training set, demonstrate the superiority of \our{}. Compared to the state-of-the-art zero-shot TTS system VALL-E, \our{} significantly improves the accuracy of synthesized speech, and demonstrates comparable or superior speaker similarity and speech naturalness on a series of subjective and objective experiments. \our{} achieves a  WER of 2.28\% on the test-clean set of LibriSpeech. Notably, \our{} works well on a wide spectrum of decoding strategies -- even greedy decoding, and still has a substantially better speech accuracy than the best of VALL-E. We further conducted ablation experiments to investigate the effects of our proposed modifications. The results indicate that the \emph{global advance} in \our{} significantly improves the model's performance, while the local advance enhances the stability of the generated output. 

\section{Related Work}

\paragraph{language modeling}

Recently, language models have garnered increasing interest in both the academic and industrial communities. Compared to models that are confined to specific tasks, language models have been proven to possess the capability to solve a wide array of tasks, shining across various domains such as text~\citep{gpt3,palm,rae2021scaling,yu2022scaling}, images~\citep{alayrac2022flamingo,tsimpoukelli2021multimodal}, and videos~\citep{yang2022zero,wang2022language}. In the audio domain, AudioLM~\citep{audiolm} trains language models on discretized audio tokens, achieving speech synthesis tasks through hierarchical prediction of these tokens. AudioGen~\citep{audiogen} employs an auto-encoding approach to extract discrete encodings of raw audio, and trains a language model conditioned on textual features for controlled audio generation. LM-VC~\citep{lmvc} employs three language models—a masked prefix language model, an external LM, and a prefix LM—to achieve zero-shot voice conversion. ~\citet{kakouros2023investigating} investigates the role of word surprisal, extracted from language models, in influencing the prosody of speech synthesized by TTS systems. For zero-shot TTS, ~\citet{VALLE} approaches TTS as a conditional language modeling task rather than a continuous signal regression. By employing discrete audio codes obtained from pre-trained neural codec, it trains a discrete audio language model, achieving improved naturalness in speech and preservation of speaker characteristics. VALL-E-X~\citep{VALLEx} extends VALL-E by utilizing source language speech and target language text as prompts when predicting the acoustic marker sequence of the target language speech. This approach supports high-quality zero-shot cross-lingual voice synthesis. These methods require only a single utterance of an unknown speaker as a prompt to generate high-quality, specified speech. 

\paragraph{speech synthesis}

Speech synthesis has long been a significant topic in the fields of artificial intelligence, natural language processing, and speech processing. Early methods were based on Statistical Parametric Speech Synthesis (SPSS)~\citep{zen2009statistical}, typically involving complex components such as text analysis models, acoustic models, and vocoders (\eg, hidden Markov model(HMM)~\citep{yoshimura1999simultaneous} based). While cost-effective in terms of data, the generated speech of SPSS still exhibited noticeable differences from natural human speech. With the advancement of modern neural networks, some work initially replaced HMMs with recurrent neural networks (RNNs) but still followed the SPSS paradigm~\citep{fan2014tts, zen2015unidirectional, valentini2016investigating}. Later, end-to-end neural TTS models were introduced, which synthesize Mel spectrograms and employ a vocoder~\citep{wavenet, waveglow} for speech synthesis~\citep{tacotron,deepvoice,fastspeech}. Some methods, utilizing techniques such as VAE~\citep{hsu2018hierarchical, lee2022bidirectional}, flow~\citep{flowtts, glowtts}, diffusion~\citep{difftts, guidedtts, gradtts}, and others~\citep{wu2022itowave}, have achieved promising performance in end-to-end speech synthesis. On the other hand, models like VALL-E~\citep{VALLE} and AudioLM~\citep{audiolm} utilize autoregressive Transformers to model discrete audio tokens, achieving great in-context learning performance. When it comes to zero-shot speech synthesis, autoregressive Transformer-based models can predict and generate audio without the need for an additional duration model, which strikes a favorable balance between efficiency and performance, and has been garnering increasing attention.

%
\begin{figure*}[t]
    \centering
    \includegraphics[width=0.9\linewidth]{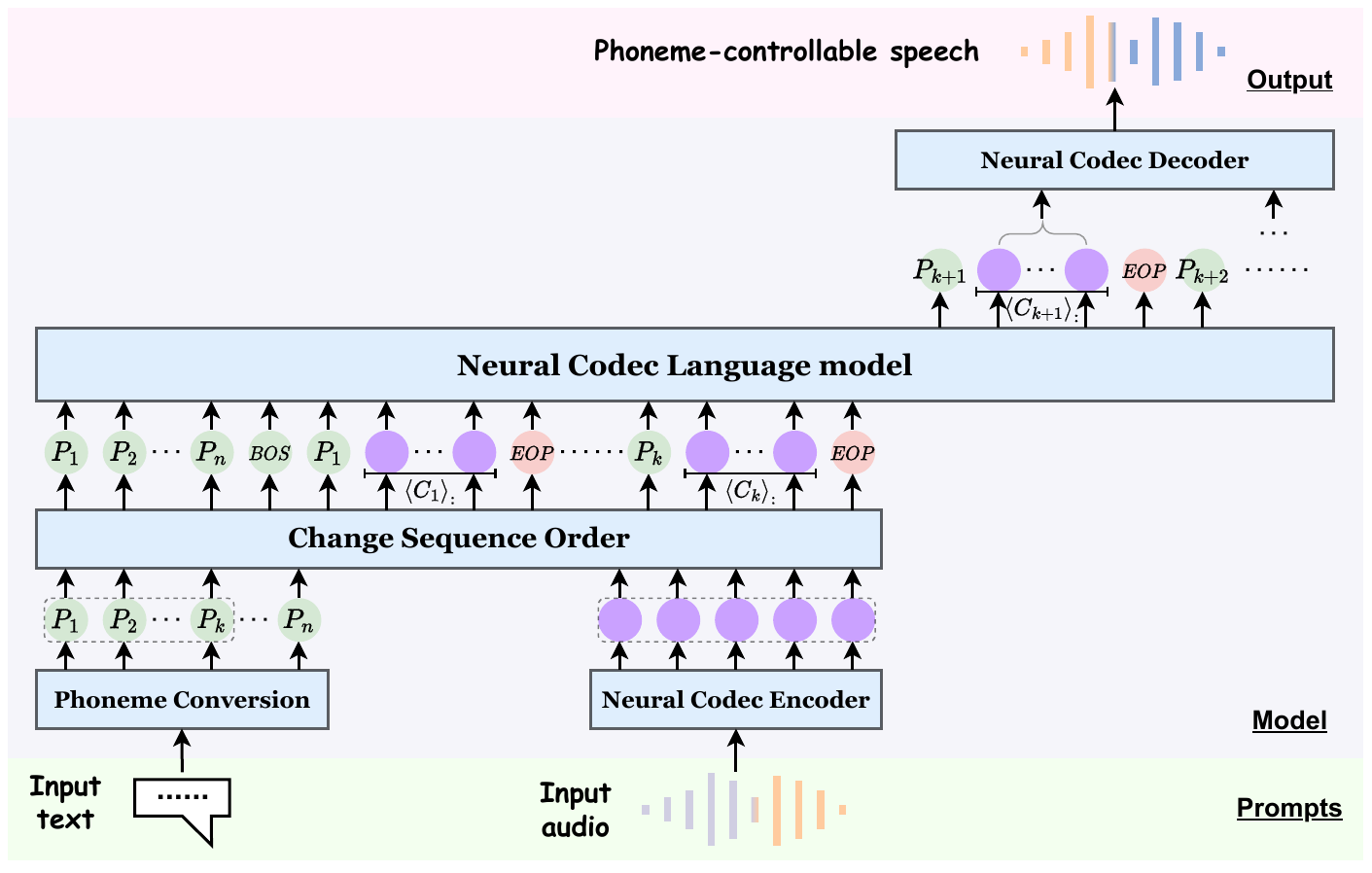}
    \caption{
        The overall architecture of \our. Input an audio prompts and text prompts, \our{} first changes  sequence order -- sandwiching each phoneme's audio $\left<\mathbf{C}_k\right>_{:}$ between the $k$-th phoneme and a \texttt{EOP} token and prepending the phoneme sequence to the beginning. By learning on the mixed sequence, \our{} can generate audio sequence of the text prompts while maintaining the acoustic and environmental conditions of the audio prompts.
    }
    \label{fig:overall}
\end{figure*}
%

\section{Method}

\subsection{Overview}

Fig.~\ref{fig:overall} demonstrates the overall architecture of \our. \our{} primarily follows a two-stage framework similar to VALL-E, considering zero-shot TTS as a conditional codec language modeling task. \our{} maps input text prompts and speech prompts into a unified vocabulary space with a text encoder and a neural codec, respectively. Different from VALL-E, an additional sequence order rearranging step is performed to the text-audio token sequence, after which,  \our{} utilizes a decoder-only language model to learn to perform conditional generation on the hybrid sequences of phoneme and audio tokens. Detailed information about the language model will be presented in Section~\ref{sec:training}. 

To obtain discrete audio representations, we employ a pre-trained neural audio codec model, EnCodec~\citep{encodec}, following VALL-E~\citep{VALLE}. EnCodec transforms 24 kHz raw waveforms into 75 Hz discrete tokens using $L$ RVQ layers. The discrete acoustic tokens have a hierarchical structure, where the first layer quantizer contains semantic information and coarse-grained acoustic contours, while subsequent $L-1$ quantizers learn fine-grained acoustic details. In our experiments, we use the same settings as VALL-E, with $L=8$. For each quantizer, we set the codebook size to 1024. In this setting, each second of the waveform is represented by $75\times 8$ discrete tokens from RVQ. 

To obtain phoneme sequences, we apply the Montreal Forced Aligner (MFA)~\citep{mfa} to the input audio and text transcriptions. Notably, MFA not only serves as a text tokenizer but also extracts alignment relationships between phonemes and the corresponding speech. The forced alignment information is essential for \our{} to \textbf{change sequence order}. In Section~\ref{sec:training}, we will provide a detailed explanation of how this information is used to construct the target sequence.

\subsection{Training: Codec Language Model}
\label{sec:training}

\our{} employs a Generalized Autoregressive Codec language model for the prediction of the first quantization layer in the EnCodec, which corresponds to capturing semantic information and coarse-grained acoustic profiles. Subsequently, a non-autoregressive language model is utilized to generate codes for the subsequent quantization layers, aimed at reconstructing fine-grained acoustic details. Specifically, given a speech corpus $\mathcal{D} = \{\mathbf{x}_i , \mathbf{y}_i \}$, where $\mathbf{x}$ represents an audio sample, and $\mathbf{y}$ is its text transcription. We utilize the EnCodec to extract the discrete representation of $\mathbf{x}$, formulated as
  \begin{equation}
    \mathbf{C}^{T\times 8} = \text{EnCodec} (\mathbf{y})
  \end{equation}
where $\mathbf{C}$ represents the two-dimensional acoustic code matrix, and $T$ is the downsampled utterance length.

We employ MFA to obtain the phoneme sequence $\mathbf{P}_{1:n}$ corresponding to the transcription $\mathbf{y}$, while also extracting forced alignment information between the audio $\mathbf{x}$ and the transcription $\mathbf{y}$:
  \begin{equation}
    \mathbf{P}_{1:n}, \bm{l}_{1:n} = \text{MFA} (\mathbf{x}, \mathbf{y})
  \end{equation}
where $n$ is the number of phonemes of the audio sample $\mathbf{x}$, and $l_i$ denotes the length of the $i$-th phoneme of the discrete audio sequence. MFA treats silence also as a kind of phoneme, so that the original audio sequence is partitioned into $n$ consecutive intervals corresponding to $n$ phonemes. Specifically, let $\left<\mathbf{C}_i\right>^{l_i\times 8}$ represent the audio sequence corresponding to the $i$-th phoneme:
  \begin{equation}
    \mathbf{C} = 
        \begin{bmatrix}
            \left<\mathbf{C}_1\right>_{1:l_1}  \\
            \left<\mathbf{C}_2\right>_{1:l_2}  \\
            \vdots  \\
            \left<\mathbf{C}_n\right>_{1:l_n}  \\
        \end{bmatrix}
  \end{equation}
and
  \begin{equation}
    \left<\mathbf{C}_k\right>_{1:l_k} = \mathbf{C}_{\sum_{i=1}^{k-1} l_i+1 : \sum_{i=1}^{k} l_i}
  \end{equation}
  
After quantization, we utilize the EnCodec decoder to reconstruct the audio waveform from the discrete acoustic sequence $\mathbf{C}$, formulated as $\hat{\mathbf{x}} \approx \text{DeCodec}(\mathbf{C})$.

For the zero-shot TTS task, the optimization objective is $\max p (\mathbf{C} |\mathbf{P}, \mathbf{\hat{C}})$, where $\mathbf{\hat{C}}$ is the acoustic prompt of the unseen speaker. We use language modeling to generate acoustic tokens for the unseen speaker, by learning on the mixed sequence composed of phonemes and codec codes, consistent with previous works~\citep{VALLE,audiopalm}.

Unlike existing approaches, \our{} does not concatenate phoneme tokens and acoustic tokens directly to form the target sequence for training the language model. Instead, \our{} \textbf{interleaves phoneme and acoustic tokens} in order to make it easier for language models to learn the alignment between audio and text. Specifically, we insert each phoneme token $P_i$ (except the \emph{silence phoneme}) into the corresponding position of the audio sequence, so that each phoneme's audio $\left<\mathbf{C}_i\right>$ is sandwiched between $P_i$ and \verb|EOP| tokens. We also prepend the phoneme sequence to the beginning of the mixed sequence, which is referred to as \emph{global advance}. In Section \ref{sec:local}, we further propose a variant sequence order with higher generation stability, named \emph{local advance}, which moves the non-acoustic tokens of the sequence several frames forward. 

%
\begin{figure}[t]
    \centering
    \includegraphics[width=1.0\linewidth]{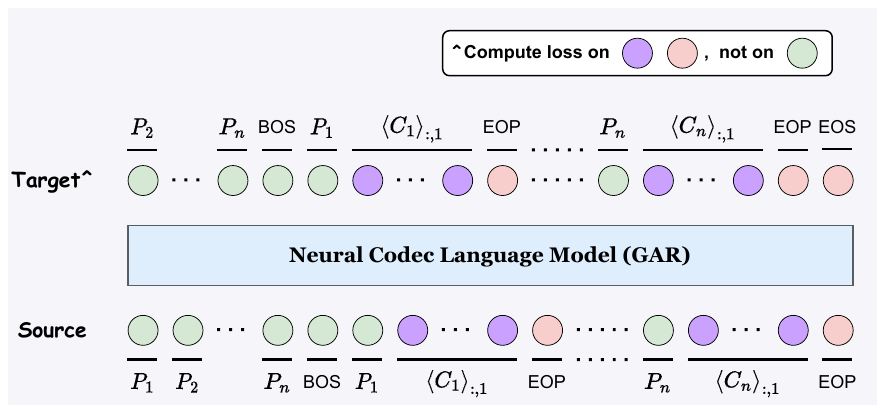}
    \caption{
        The illustration of Generalized Autoregressive language model of \our. 
    }
    \label{fig:ar}
\end{figure}
%

\subsubsection{Generalized Autoregressive (GAR) Codec Language Model}
\label{sec:gar}

As shown in Figure \ref{fig:ar}, \our{} first constructs a hybrid sequence $\mathbf{H}_{:,1}$ of acoustic and phoneme tokens, structured as:
\begin{center}
  $P_1, P_2, \ldots{}, P_n, \texttt{BOS}, P_1, \left<\mathbf{C}_1\right>_{:,1}, \texttt{EOP},$\\ 
  $P_2, \left<\mathbf{C}_2\right>_{:,1}, \texttt{EOP}, \ldots{}, P_n, \left<\mathbf{C}_n\right>_{:,1}, \texttt{EOP}, \texttt{EOS}$
\end{center}
It is worth noting that the MFA (Montreal Forced Aligner) treats silence as a distinct phoneme, whereas our phoneme sequence $\mathbf{P}$ exclusively comprises phonemes other than silence. To clarify, we retain the acoustic component associated with silence but do not sandwich it with an \verb|EOP| and a specific silence phoneme, nor do we use a silence phoneme in the \emph{global advance} part. 

We design a GAR language model to learn the continuation task on the aforementioned hybrid sequence, to generate the discrete acoustic code sequence $\mathbf{C}_{:,1}$. The GAR model consists of multiple Transformer decoder layers~\citep{transformer}. After training, it can generate discrete audio codes for a specified text prompt and acoustic prompt. GAR is also responsible for predicting \verb|EOP| and \verb|EOS| to indicate the conclusion of a phoneme and the entire sentence, respectively.

The optimization of GAR is achieved by maximizing the likelihood of the acoustic part $\mathbf{C}_{:,1}$ of the hybrid sequence $\mathbf{H}_{:,1}$, as well as the special \verb|EOP| and \verb|EOS| tokens. Under forward factorization, this process is formulated as:
  \begin{equation}
  \begin{aligned}
    &\max\limits_{\theta_{GAR}} \ \; 
    \log p ( \mathbf{\tilde{C}}_{:,1} \,|\, \mathbf{P} ; \theta_{GAR} ) \\
    = 
      &\sum_{i=1}^{n}
      \sum_{t=0}^{l_{i}} 
      \log p \bigl(
        \bigl<\mathbf{\tilde{C}}_{i}\bigr>_{t,1} | 
        \bigl<\mathbf{\tilde{C}}_{i}\bigr>_{<t,1}, 
        \bigl<\mathbf{\tilde{C}}_{<i}\bigr>_{:,1}, \\
        &\mathbf{P} ; \theta_{GAR}
      \bigr) \\ 
    =
      &\sum_{\substack{t=0 \\ \mathbf{H}_{t,1} \neq \texttt{BOS} \\ \mathbf{H}_{t,1}  \notin \{\mathbf{P}\}}}^{T_{\bm{H}}} \log p \left( 
        \mathbf{H}_{t,1} \,|\, \mathbf{H}_{<t,1} ; \theta_{GAR}
      \right)
  \end{aligned}
  \end{equation}
where $\mathbf{H}$ has a size of $T_{\bm{H}} \times 8$, $\{\mathbf{P}\}$ denotes the phoneme set, $\bigl<\mathbf{\tilde{C}}_{i}\bigr>$ is the concatenation of $\left<\mathbf{C}_i\right>$ along with its broadcast trailing \verb|EOP| and/or \verb|EOS| tokens, $\mathbf{\tilde{C}}$ is then the concatenation of $\left<\mathbf{C}_i\right>$,  and $\theta_{GAR}$ represents neural network parameters of GAR model. The factorization of the training objective naturally encapsulates the core intuition of the GAR model: GAR generates the audio sequence phoneme-by-phoneme. GAR produces maximum likelihood predictions for each phoneme token successively, indicating the end of generating a specified phoneme by predicting \verb|EOP|. Through \emph{global advancement}, GAR can directly infer the next phoneme to be generated without relying on network predictions. After the prediction for the last phoneme is completed, GAR stops the generation process by predicting \verb|EOS|. The generated sequence by GAR is \textbf{self-aligned}, as it can instantly know the corresponding position of any generated acoustic token in relation to the phoneme prompt.

During training, we apply a bidirectional mask to the phoneme sequence before the \verb|BOS| in the hybrid sequence, while a unidirectional mask is used for the part after \verb|BOS|. We frame the training as a next-token-prediction language modeling task on the hybrid sequence. However, it's important to note that the model does not predict phonemes (or \verb|BOS|). In other words, as shown in Figure~\ref{fig:ar}, we only compute loss when the token to be predicted is not a phoneme (or \verb|BOS|). During inference, whenever the model predicts an \verb|EOP| for a phoneme, the next phoneme token is directly appended to the end of the sequence, which will be further discussed in Section~\ref{sec:exp}.

\subsubsection{Non-Autoregressive (NAR) Codec Language Model}

In the second stage, the NAR language model is employed to predict the codes from the second to the last quantization layers in parallel. The input-output sequence construction of the NAR model follows the same pattern as used in the GAR model discussed in Section~\ref{sec:gar}. Specifically, the $i$-th column $\mathbf{H}_{:, i}$ of the hybrid sequence matrix $\mathbf{H}$ is structured as:
\begin{center}
  $P_1, P_2, \ldots{}, P_n, \verb|BOS|, P_1, \left<\mathbf{C}_1\right>_{:,i}, \verb|EOP|, P_2, \left<\mathbf{C}_2\right>_{:,i},$ \\
  $\verb|EOP|, \ldots{}, P_n, \left<\mathbf{C}_n\right>_{:,i}, \verb|EOP|, \verb|EOS|$
\end{center}
And in practice if $P_i$ represents the silence, $\mathbf{C}_{:,i}$ will not be sandwiched by $P_i$ and \verb|EOP|.

The NAR model takes the previously generated hybrid sequence of the previous $j-1$ layers as input and predicts the codes of the $j$-th layer in parallel, formulated as:
  \begin{equation}
  \begin{aligned}
    &\max\limits_{\theta_{NAR}} \ \;
    \sum_{j=2}^{8}
    \log p ( \mathbf{C}_{:,j} \,|\, \mathbf{H}_{:,<j}, \mathbf{P} ; \theta_{NAR} ) \\
    = 
      &\sum_{j=2}^{8}
      \sum_{\substack{t=0 \\ \mathbf{H}_{t,j} \in \{\mathbf{C}_{:,j}\}}}^{T_{\bm{H}}}
        \log p ( \mathbf{H}_{t,j} \,|\, \mathbf{H}_{:,<j}, \mathbf{P} ; \theta_{NAR} )
  \end{aligned}
  \end{equation}
where $\{\mathbf{C}_{:,j}\}$ denotes the acoustic token set of the $j$-th quantizer. In this formulation, 
The embeddings of tokens from the previous $j-1$ quantizers are summed up to feed the NAR model to predict the $j$-th layer. Intuitively, both the GAR and NAR model of \our{} compute the loss on the acoustic tokens of the target sequence, and GAR additionally computes loss for \verb|EOP| and \verb|EOS|.

%
\begin{figure}[t]
    \centering
    \includegraphics[width=1.0\linewidth]{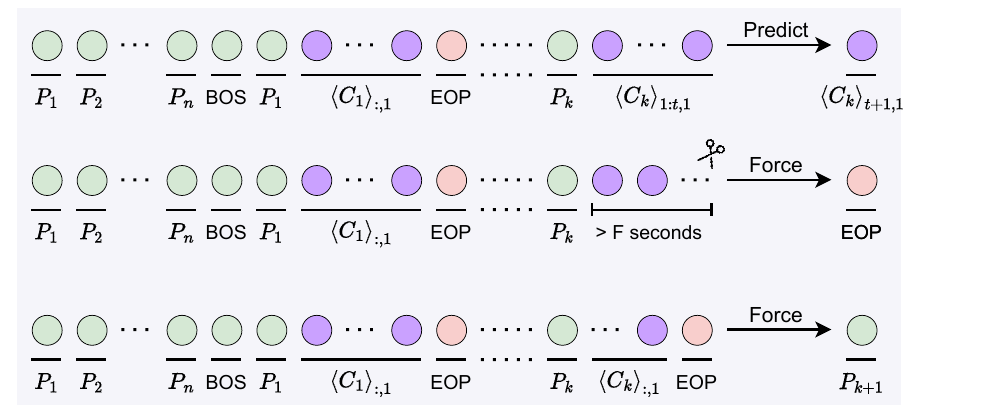}
    \caption{
        The illustration of the inference process of \our.
    }
    \label{fig:inference}
\end{figure}
%

\subsection{Inference}

\our{} can use a short clip of speech from an unseen speaker as an acoustic prompt to synthesize speech for a specified text prompt. Figure \ref{fig:inference} illustrates the inference process of the GAR model. While VALL-E may get stuck in an infinite loop during inference, resulting in the synthesis of either \emph{infinite silence} or repetitive pronunciation~\citep{VALLE}, \our{} is capable of generating \verb|EOP| and promptly truncating abnormally long phonemes. Following an \verb|EOP|, we can directly append the next phoneme token to the end of the generated sequence, ensuring the proper generation of speech without abnormal pauses or repetitions. For the GAR model, we employ a sampling-based decoding strategy, whereas for the NAR model, we use a greedy decoding approach to strike a balance between efficiency and performance. 

%
\begin{figure}[t]
    \centering
    \includegraphics[width=1.0\linewidth]{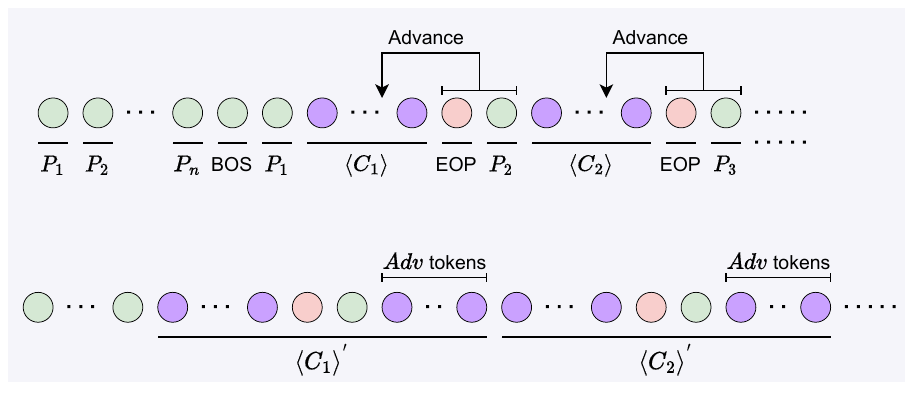}
    \caption{
        \emph{Local advance}. A phoneme can locally have access to information about the next phoneme token advanced by $Adv$ frames, allowing it to anticipate the upcoming phoneme token's characteristics.
    }
    \label{fig:advance}
\end{figure}
%

\subsection{Local Advance}
\label{sec:local}

One intuition is that the pronunciation of a phoneme is strongly related to the pronunciation of the phonemes just before and after it. However, due to the autoregressive nature of the GAR  model, an acoustic token cannot attend to the following phoneme tokens, even though we can leverage the transformer's ability to model long-term dependencies through \emph{global advance} to provide complete context for the acoustic token generation. To further harness the powerful capability of the transformer in modeling local dependencies, \our{} introduces an additional change in the sequence order based on Section~\ref{sec:training}. Specifically, we move the phoneme token and the \verb|EOP| token ahead by a few frames, referred to as \emph{local advance}. 

\section{Experiments}
\label{sec:exp}

\subsection{Experimental Setup}

\paragraph{Data \& Tasks:}
\label{sec:data}
We trained \our{} using the Librispeech~\citep{librispeech} 960h training dataset. We utilized Montreal Forced Aligner (MFA)~\citep{mfa} to obtain forced alignment information for the audio-transcription pairs. Sentences with unrecognized or unknown phones by MFA were excluded. The open-source 24kHz checkpoint~\footnote{\url{https://github.com/facebookresearch/encodec}} of EnCodec\citep{encodec} was used as the codec to generate discrete acoustic tokens. The LibriSpeech training data was upsampled to 24 kHz before feeding it into EnCodec. 

In evaluating the model, two zero-shot TTS tasks were considered. 
For the zero-shot TTS continuation task, we adhered to methodologies established by previous works~\citep{VALLE, voicebox, speechx}, selecting examples ranging from 4 seconds to 10 seconds from the LibriSpeech test-clean dataset as our test set. 
In this task, we used the complete phoneme transcription as the text prompt and the first 3 seconds of the test audio sample as the acoustic prompt. The model was required to generate continuations.

For the zero-shot TTS cross-speaker task, we designed a hard case set comprising 100 hard sentences, as outlined in the demo page \url{}. These sentences included challenging phonetic patterns, alliteration, and unusual (abnormal) combinations of words that might pose difficulties for a TTS system to generate natural-sounding speech. 
In this case, we randomly picked 3-second sentences from the LibriSpeech test-clean subset as the acoustic prompt. We then concatenated the transcription of this segment and the target phoneme sequence in the hard case set to form the text prompt. The model was tasked with cloning the voice of the speaker to say the specified target text in the hard case set.

\paragraph{Training configuration:}

For both GAR and NAR models, we stacked 12 Transformer decoder layers with 
 an embedding dimension of 1024, a hidden state dimension of 1024, and a feed-forward layer dimension of 4096. All models were trained in parallel using 8 NVIDIA Tesla V100 GPUs with a batch size of 16384 tokens for GAR and 12288 tokens for NAR per GPU, respectively, learning a total of 320k steps. We used the AdamW optimizer with $\beta_1=0.9$, $\beta_2=0.999$, $\epsilon=10^{-9}$. We employed an inverse-sqrt learning rate scheduler with warm-up. For the first $32000$ updates, we linearly increased the learning rate from $10^{-7}$ to a peak of $5 \times 10^{-4}$. The weight decay was 0.01. 

\paragraph{Baseline:}

In our research, we benchmarked the performance of zero-shot speech synthesis against VALL-E \citep{VALLE}. This system was originally trained on a substantial 60k hours of audio from the Librilight dataset \citep{librilight}. To ensure a rigorous evaluation, we reproduced the VALL-E model and adapted it to train on the LibriSpeech 960h dataset. We also adjusted the model dimensions and the number of layers to match the parameter settings of \our{} and VALL-E. Both GAR (or AR) and NAR models of VALL-E and \our{} have 154.3M parameters.
Moreover, to mitigate any potential bias introduced by the audio codec, we pre-processed the authentic speech samples using EnCodec's encoder and decoder. We also include the result for Encodec reconstructed speech for reference, denoted as Ground-Truth Encodec.

\paragraph{Evaluation Metrics:}

We evaluated our system with several objective metrics. Speaker similarity (SPK) and WER served as our primary measures. SPK was assessed using the fine-tuned WavLM-TDNN model\footnote{\url{https://huggingface.co/microsoft/wavlm-base-plus-sv}}~\citep{wavlm}, scoring similarity on a scale of -1 to 1, with values above 0.86 indicate the same speaker identity (This value comes from the release model card page). The WER was determined by comparing the synthesized speech to the original text using the Conformer-Transducer model\footnote{\url{https://huggingface.co/nvidia/stt_en_conformer_transducer_xlarge}}~\citep{conformer}. 

In addition to these standard metrics, we introduced two novel measures: INF\% and CUT\%. INF\% quantified the frequency of generating infinitely long audio, indicative of a failure in synthesis. It is used to measure the likelihood of the model falling into abnormal repetition (such as infinite silence). A higher INF\% indicates poorer stability in the generated output of the model.
In the practical implementation, INF\% referred to the proportion of sentences for which generation was not stopped when the length of the generated audio reached twice the original, serving as a proxy for infinite generation. 
On the other hand, as discussed in the previous session, the design of \our{} enables the control of the duration for each phoneme during inference, thus avoiding the synthesis failure. In our experiments, we forcibly truncate the synthesis of phonemes with a length greater than 0.4 seconds. CUT\% is used to measure the frequency of forced cuts of phonemes in synthesis by \our{}. 
For each objective metric, we reported average values over three experimental runs with different random seeds.

For subjective analysis, we relied on the mean opinion score (MOS). 30 test samples were chosen for this purpose, with each sample being evaluated by at least 15 listeners for aspects like naturalness and speaker similarity. The comparative mean option score (CMOS) and the similarity mean option score (SMOS) were the key subjective metrics used. SMOS was rated on a 1 to 5 scale, in 0.5-point increments, to gauge speaker similarity, while CMOS, ranging from -1 to 1, assessed the overall naturalness and quality of the synthesized speech against the baseline.

\begin{table}[t]
    \small
    \centering
    \caption{
        Subjective and Objective performance comparison between ELLA-V and VALL-E on zero-shot TTS continuation task. $^\dagger$ indicates that ground-truth audios were passed through the encoder and decoder of Encodec to evaluate the influence of neural audio codec. 
    }
    \resizebox{0.49\textwidth}{!}{
        \begin{tabular}{l|cccc}
            \toprule
            Models & WER(\%) ($\downarrow$) & SPK ($\uparrow$) & CMOS & SMOS \\
            
            \midrule
            Ground Truth & 1.41 & 0.923 & 0.29 & 4.39\\
            Ground Truth-Encodec$^\dagger$ & 1.62 & 0.913 & 0.22 & 4.33\\
            
            \midrule[0.15pt]
            VALL-E & 5.00 & 0.868 & 0.00 & \textbf{3.56}\\
            
            \our \textit{(ours)} & \textbf{2.28} & \textbf{0.870} & \textbf{0.10} & \textbf{3.56} \\
            
            \bottomrule
        \end{tabular}
    }
    \label{tab:results}
\end{table}

\subsection{Results}
\paragraph{Zero-Shot TTS continuation task.}
We present the evaluation results in Table~\ref{tab:results}, where a comparison between \our{} and VALL-E is shown.
First, regarding speaker similarity, both subjective (SMOS) and objective (SPK) results indicate that \our{} and VALL-E performed similarly, which can be attributed to their shared backbone approach, combining (G)AR and NAR.
Meanwhile, CMOS testing shows that \our{} achieved a +0.10 score, demonstrating a higher generation quality (i.e., naturalness) compared to VALL-E.
Additionally, WERs calculated between the recognized text of synthesized audio and the ground-truth text show that \our{} is significantly better than VALL-E (2.28 versus 5.00). This underscores \our{}'s enhanced capability in synthesizing higher-quality and more robust speech.  
Overall, \our{} substantially improved the synthesis accuracy and robustness of the language model-based TTS framework without affecting the naturalness and speaker similarity. 
This conclusion is not only corroborated by this easy continuation task, but also validated via the challenging synthesis sets in the subsequent section.

\begin{table}[t]
    \small
    \centering
    \caption{
        WER comparison between \our{} and VALL-E on 100 particularly hard synthesis cases. Sub, Del, and Ins refer to Substitution, Deletion, and Insertion error rates, respectively.
    }
    {\footnotesize
        \begin{tabular}{l|cccc}
            \toprule
            \textbf{Models} & WER(\%) & Sub(\%) & Del(\%) & Ins(\%) \\
            
            \midrule
            VALL-E & 28.39 & 17.79 & 5.36 & 5.24 \\
            
            \our & \textbf{12.79} & \textbf{7.76} & \textbf{3.40} & \textbf{1.63} \\
             
            \bottomrule
        \end{tabular}
    }
    \label{tab:hard}
\end{table}

\paragraph{Zero-shot TTS cross-speaker task on hard cases.}
\label{sec:hard}
VALL-E utilized a traditional AR model that frequently resulted in alignment errors, including repetitions, transpositions, and omissions, particularly in more challenging synthesis cases (see Section \ref{sec:data} for details of the challenging synthesis set). 
Table \ref{tab:hard} presents the WER comparison of VALL-E and \our{} on the 100 particularly hard synthesis sentences.
In contrast to VALL-E, \our{} demonstrates markedly lower WER, signifying its enhanced robustness. This substantial reduction in errors translates to more accurate and reliable voice synthesis applications, significantly improving user experience in real-world scenarios.

Regarding VALL-E's tendency to fall into infinite silence, an intuitive explanation is that the silence patterns in the training data are relatively simple and many of them are repetitive. In this case, a traditional language model is prone to overfitting to these patterns. During testing, when the model encounters silence, it assigns a high probability to silence. This leads to issues such as beam search, which is based on maximum likelihood, getting stuck in a loop. However, \our{} does not face this problem.

%
\begin{figure*}[t]
    \centering
    \begin{subfigure}{0.49\textwidth}
        \includegraphics[width=\linewidth]{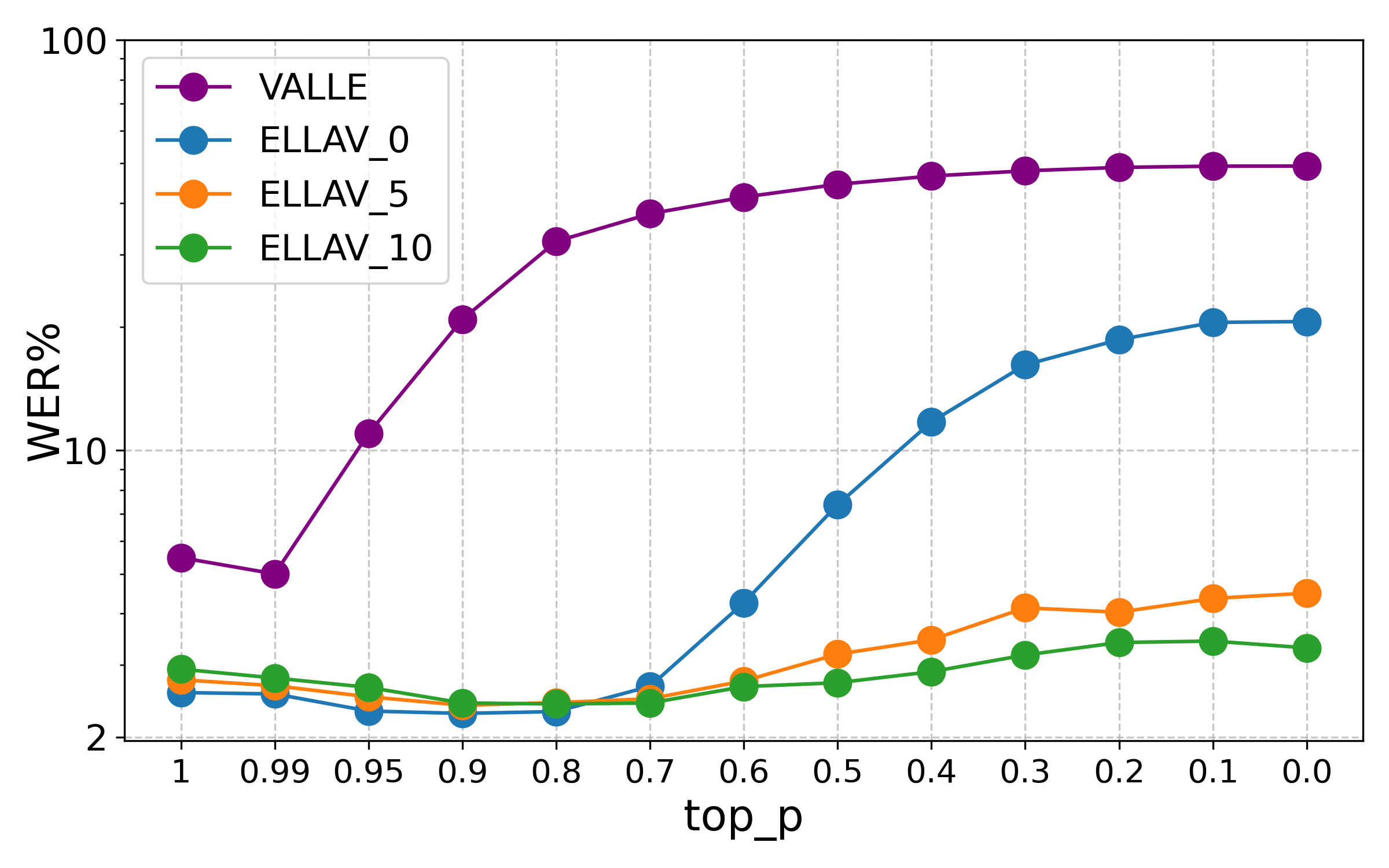}
        \label{fig:result2c}
    \end{subfigure}
    \hfill
    \begin{subfigure}{0.49\textwidth}
        \includegraphics[width=\linewidth]{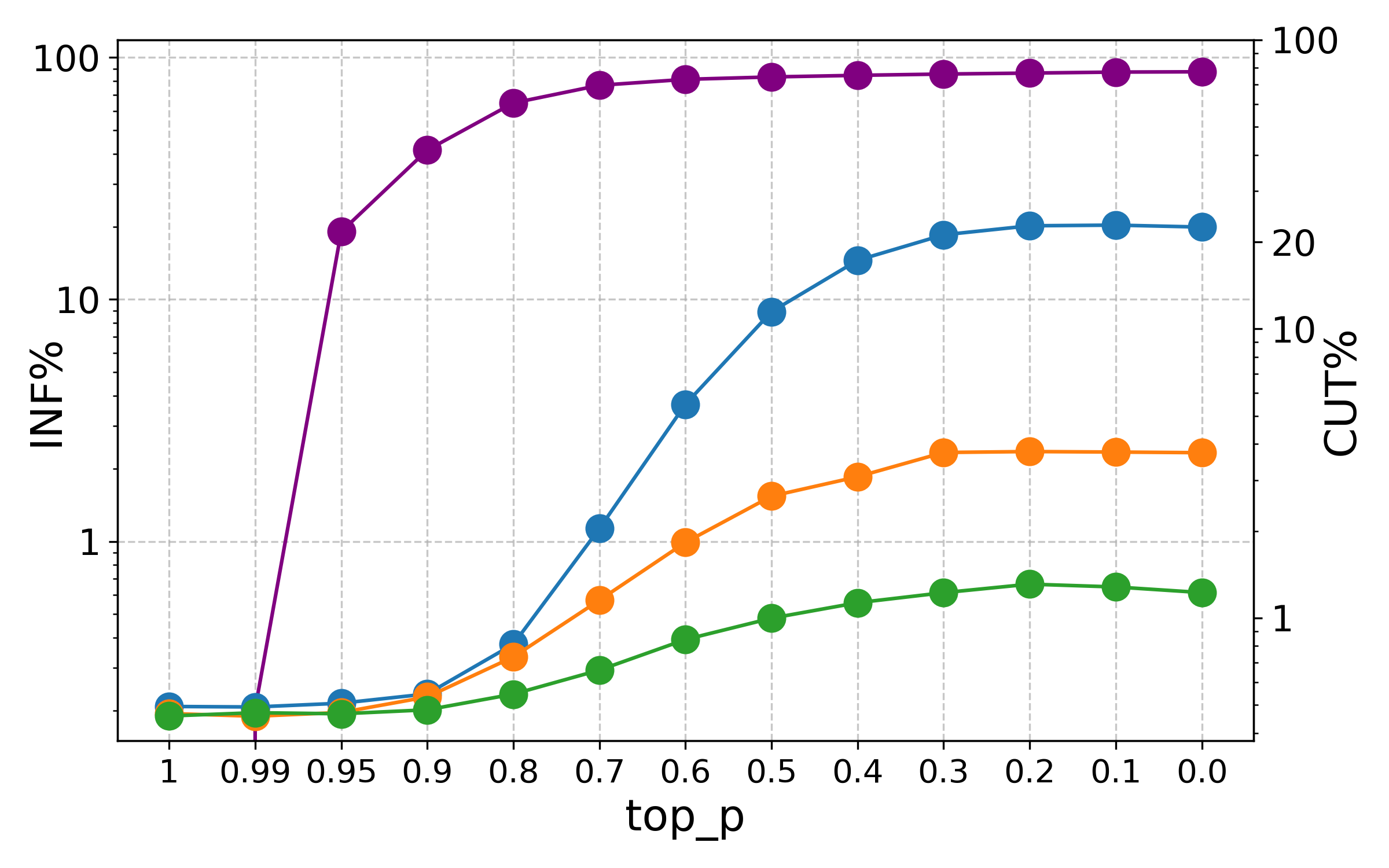}
        \label{fig:result3c}
    \end{subfigure}
    \caption{
        Ablations on decoding strategies. The figures demonstrate the trends of three metrics, INF (for VALL-E), CUT (for \our), and WER (for both), with respect to the variations in top\_p in nuclear sampling.
    }
    \label{fig:results}
\end{figure*}

\paragraph{Analysis of Decoding Strategies.} 
To demonstrate the stability of \our{} under different decoding strategies, we conducted an ablation study, testing the decoding performance with different top-$p$ values for nuclear sampling, by varying $p \in \{1, 0.99, 0.95, 0.9, 0.8, 0.7, 0.6, 0.5, 0.4, 0.3, 0.2,$ $0.1, 0.0(\text{greedy})\}$. The results are shown in Figure \ref{fig:results}. 
We can observe that as top\_$p$ decreases, the accuracy of VALL-E's synthesized speech significantly decreases. At this point, VALL-E is more prone to generating a large number of overfit silence tokens, leading to a significant increase in INF\%. And compared to VALL-E, the audio synthesized by \our{} is less sensitive to rate changes in the top\_$p$ sampling strategy, whose WER consistently outperforms VALL-E. When the local advance is set to 5 or 10 tokens, the generated audio exhibits significant stronger robustness. On the other hand, as shown in Figure \ref{fig:results} (right), as top\_$p$ decreases, VALL-E tends to get stuck in infinite loops of failed generation, while the generation of \our{} remains significantly stable. Moreover, \our{} can promptly handle (truncate) the synthesis of exceptional phonemes, resulting in significantly higher robustness.

\begin{table}[t]
    \small
    \centering
    \caption{
       The ablation study to investigate the impact of global and local phoneme information.
    }
    {\footnotesize
        \begin{tabular}{l|cc}
            \toprule
            Models & WER(\%) ($\downarrow$) & SPK ($\uparrow$) \\
            \midrule
            
            VALL-E & 5.00 & 0.868 \\
            \midrule[0.1pt]
            
            \our & \textbf{2.28} & \textbf{0.870} \\
            
            \our-noglobal & 5.00 & 0.859 \\
             
            \our-nophn & 3.51 & 0.868 \\
            
            \bottomrule
        \end{tabular}
    }
    \label{tab:ablation}
\end{table}

\paragraph{Ablation Study.}
In this paragraph, we conduct ablation experiments. (1) To investigate the impact of global phoneme information on synthesized speech, we removed the global phoneme sequence at the beginning of the trained sequence (abbr. \textbf{\our{}}-noglobal). (2) To investigate whether it is necessary to provide the specific phoneme token before its corresponding acoustic tokens during both training and inference, rather than just using the \verb|EOP| separator, we removed all phoneme tokens following \verb|BOS| in the mixed sequence (abbr. \textbf{\our}-nophn). 
The experimental results are shown in Table \ref{tab:ablation}. 
It is observed that the accuracy of synthesized speech significantly deteriorated either when global phoneme tokens were not used or when local phoneme tokens were disabled within the hybrid sequence.
It is also notable that even in the absence of global advance (i.e., in the \our{}-noglobal configuration), the SPK and WER of the synthesized audio were comparable to those of VALL-E. These findings indicate the importance of both local and global information in achieving more accurate synthesized audios, meanwhile, combining both of them potentially leads to further enhancements in accuracy.

\section{Conclusion}

In this paper, we introduce \our{}, a simple and efficient two-stage zero-shot TTS framework based on language modeling. By learning interleaved sequences of acoustic and text tokens, our proposed GAR model can provide fine-grained control over synthesized audio at the phoneme level and can better leverage local dependencies to predict the pronunciation of the current phoneme. Experimental results demonstrate that \our{} achieves higher accuracy and more stable results under different threshold top-$p$ for nuclear sampling. We aspire for this work to advance research in enhancing the robustness of speech generation.

\bibliography{custom}
\bibliographystyle{acl_natbib}

\end{document}